\documentclass[10pt,twocolumn,letterpaper]{article}
\usepackage{iccv}
\usepackage{times}
\usepackage{epsfig}
\usepackage{graphicx}
\usepackage{amsmath}
\usepackage{amssymb}
\usepackage{booktabs}
\usepackage{multirow}
\usepackage{multicol}

\usepackage[pagebackref=true,breaklinks=true,letterpaper=true,colorlinks,bookmarks=false]{hyperref}

\usepackage[capitalize]{cleveref}
\crefname{section}{Sec.}{Secs.}
\Crefname{section}{Section}{Sections}
\Crefname{table}{Table}{Tables}
\crefname{table}{Tab.}{Tabs.}

\usepackage{amssymb}% http://ctan.org/pkg/amssymb
\usepackage{pifont}% http://ctan.org/pkg/pifont

\iccvfinalcopy % *** Uncomment this line for the final submission

\begin{document}

%%%%%%%%% TITLE
\title{CAT: Collaborative Adversarial Training}

\author{
Xingbin Liu$^1$ \quad Huafeng Kuang$^1$ \quad Xianming Lin$^1$ \quad Yongjian Wu$^2$ \quad Rongrong Ji$^{1}$
\\
$^{1}$ Media Analytics and Computing Lab, Department of Artificial Intelligence,\\
School of Informatics, Xiamen University, 361005, China.\\
$^2$ Tencent Youtu Lab, Shanghai, China\\
}

\maketitle

\begin{abstract}
Adversarial training can improve the robustness of neural networks.
Previous methods focus on a single adversarial training strategy and do not consider the model property trained by different strategies. 
By revisiting the previous methods, we find different adversarial training methods have distinct robustness for sample instances.
For example, a sample instance can be correctly classified by a model trained using standard adversarial training (AT) but not by a model trained using TRADES, and vice versa.
Based on this observation, we propose a collaborative adversarial training framework to improve the robustness of neural networks.
Specifically, we use different adversarial training methods to train robust models and let models interact with their knowledge during the training process.
\textbf{C}ollaborative \textbf{A}dversarial \textbf{T}raining (\textbf{CAT}) can improve both robustness and accuracy. 
Extensive experiments on various networks and datasets validate the effectiveness of our method.
CAT achieves state-of-the-art adversarial robustness without using any additional data on CIFAR-10 under the Auto-Attack benchmark.
Code is available at \url{https://github.com/liuxingbin/CAT}.
\end{abstract}

\section{Introduction}
With the development of deep learning, Deep Neural Networks (DNNs) have been applied to various visual tasks, such as image classification~\cite{he2016deep}, object detection~\cite{redmon2016you}, semantic segmentation~\cite{pal1993review}, etc. 
And state-of-the-art performance has been obtained. 
But recent research has found that DNNs are vulnerable to adversarial perturbations~\cite{goodfellow2014explaining}. 
A finely crafted adversarial perturbation by a malicious agent can easily fool the neural network. 
This phenomenon raises security concerns about the deployment of neural networks in security-critical areas such as Autonomous driving~\cite{chen2019model} and medical diagnostics~\cite{kong2017cancer}.

To cope with the vulnerability of DNNs, different types of methods have been proposed to improve the robustness of neural networks, including adversarial training~\cite{madry2017towards}, defensive distillation~\cite{papernot2016distillation}, feature denoising~\cite{xie2019feature} and neural network pruning~\cite{madaan2020adversarial}. 
Among them, Adversarial Training (AT) is the most effective method to improve adversarial robustness. 
AT can be regarded as a type of data augmentation strategy that trains neural networks based on adversarial examples crafted from natural examples. 
AT is usually formulated as a min-maximization problem, where the inner maximization generates adversarial examples, while the outer minimization optimizes the parameters of the model based on the adversarial examples generated by the inner maximization process.

\begin{figure*}[t]
    \begin{center}
    \includegraphics[width=0.9\textwidth]{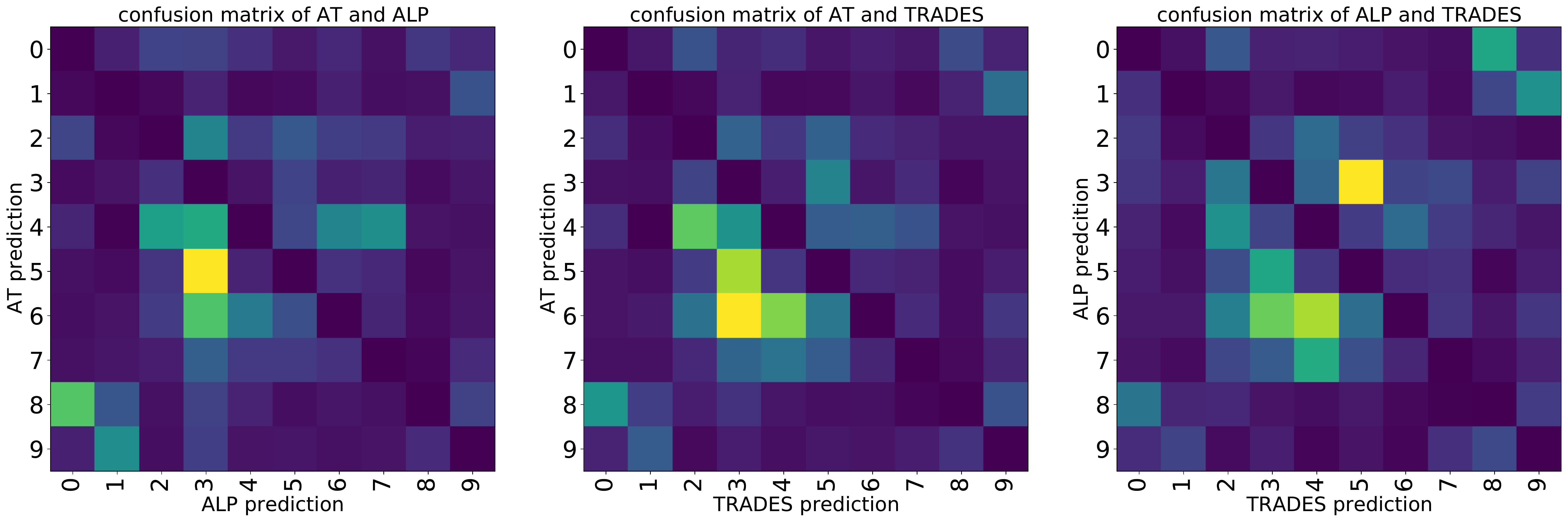}
    \end{center}
    \caption{Confusion matrices of models trained by different methods with ResNet-18 on the CIFAR-10 test dataset (better seen with color).
    We set the diagonal value as 0 for better illustration.
    Confusion exists in models trained by any two methods, especially for blocks from class 3 to class 7.
    The value of the prediction discrepancy is 18.98\%, 22.54\%, and 21.05\% respectively.}
    \label{fig:motivation}
\end{figure*}

The previous methods have focused on how to improve the model's adversarial accuracy, focusing only on the numerical improvement, but not on the characteristics of the different methods. 
By revisiting the process of adversarial defense, a question comes to our mind: \textit{Do models trained by different adversarial training methods perform the same for sample instances?}
We analyzed different adversarial training methods and found that classification confusion exists in models adversarially trained by different methods, as illustrated in \cref{fig:motivation}.
Specifically, taking AT~\cite{madry2017towards} and TRADES~\cite{zhang2019theoretically} as an example, for the same adversarial example, the network trained by AT can classify correctly, while the network trained by TRADES misclassifies, and vice versa.
A conclusion can be drawn that although AT and TRADES have similar numerical adversarial accuracy, they behave differently for sample instances, \ie, different knowledge is mastered by models trained by different methods.
This raises a question:
\begin{quote}
    \begin{center}
    \textit{Do two networks learn better if they collaborate?}
    \end{center}
\end{quote}

Based on this observation, we propose a \textbf{C}ollaborative \textbf{A}dversarial \textbf{T}raining (\textbf{CAT}) framework to improve the robustness of neural networks.
Our framework is shown in \cref{fig:framework}.
Specifically, we simultaneously train two deep neural networks separately using different adversarial training methods. 
At the same time, the adversarial examples generated by this network are input to the peer network to obtain the corresponding logit, which is used to guide the learning of this network together with its own adversarial training objective function. 
We expect to improve the robustness of the neural network by allowing peers to learn from each other in this collaborative learning way. 
Extensive experiments on different neural networks (VGG, MobileNet, ResNet) and different datasets (CIFAR, Tiny-ImageNet) demonstrate the effectiveness of our approach. 
CAT achieved new state-of-the-art robustness without any additional synthetic or real data on CIFAR-10 under the Auto-Attack benchmark.
Furthermore, we provide property analysis for CAT to get a better understanding.

In summary, our contribution is threefold as follows.
\begin{itemize}
    \item We find that the models obtained using different adversarial training methods have different representations for individual sample instances.
    \item We propose a novel adversarial training framework: Collaborative Adversarial Training. CAT simultaneously trains neural networks from scratch using different adversarial training methods and allows them to collaborate to improve the robustness of the model.
    \item We have conducted extensive experiments on a variety of datasets and networks, and evaluated them on state-of-the-art attacks. We demonstrate that CAT can substantially improve the robustness of neural networks and obtain new state-of-the-art performance without any additional data.
\end{itemize}

\section{Related Work}

\subsection{Adversarial Attack}
Since Szegedy~\cite{szegedy2013intriguing} discovers that DNNs are vulnerable to adversarial examples, a large number of works have been proposed to craft adversarial examples. 
Based on the accessibility to the knowledge of the target model, it can be divided into white-box attacks and black-box attacks. 
White-box attacks craft adversarial examples based on the knowledge of the target model, while black-box attacks are agnostic to the knowledge of the target model.

\textbf{White-box Attack:}
Goodfellow~\cite{goodfellow2014explaining} proposes FGSM to efficiently craft adversarial examples, which can be generated in just one step. 
Madry proposes PGD to generate adversarial examples, which is the most efficient way of using the first-order information of the network. 
MI-FGSM~\cite{dong2018boosting} combines momentum into the iterative process to help the model escape from local optimal points. 
And the adversarial examples generated by this method are also more transferable.
Boundary-based attacks such as deepfool~\cite{moosavi2016deepfool} and CW~\cite{carlini2017towards} also make the model more challenging.
Recently, the ensemble approach of diverse attack methods (Auto-Attack), consisting of APGD-CE~\cite{croce2020reliable}, APGD-DLR~\cite{croce2020reliable}, FAB~\cite{croce2020minimally} and Square Attack~\cite{andriushchenko2020square}, become a benchmark for testing model robustness.

\textbf{Black-box Attack:}
Block-box attacks can be categorized into transfer-based and query-based attacks.
Transfer-based methods attack the target model by using the transferability of adversarial examples, \ie, the adversarial examples generated on the surrogate model can be transferred to fool the target model. 
There are many ways to explore the transferability of adversarial examples for black-box attacks. 
Dong~\cite{dong2018boosting} combines momentum with an iterative approach to obtain better transferability.
Scale-invariance~\cite{lin2019nesterov} boosts the transferability of adversarial examples by transforming the inputs on multiple scales.
Square Attack~\cite{andriushchenko2020square} approximates model's decision boundary based on a randomized search scheme to be the most efficient query-based attack method.

\subsection{Adversarial Robustness}
Adversarial attacks present a significant threat to DNNs.
For this reason, many methods have been proposed to defend against adversarial examples, including denoising~\cite{xie2019feature}, adversarial training~\cite{madry2017towards}, data aumentation~\cite{aug}, and input purification~\cite{naseer2020self}. 
ANP~\cite{madaan2020adversarial} finds the vulnerability of latent features and uses pruning to improve robustness. 
Madry uses PGD to generate adversarial examples for adversarial training, which is also the most effective way to defend against adversarial examples. 
A large body of work uses new regularization or objective functions to improve the effectiveness of standard adversarial training. 
Adversarial logit pairing~\cite{kannan2018adversarial} improves robustness by encouraging the logits of normal and adversarial examples to be closer together. 
TRADES~\cite{zhang2019theoretically} uses KL divergence to regularize the output of adversarial and pure examples.

\subsection{Knowledge Distillation}
Knowledge distillation (KD) is commonly used for model compression and was first used by Hinton~\cite{hinton2015distilling} to distill knowledge from a well-trained teacher network to a student network. 
KD can significantly improve the accuracy of student models. 
There have been many later works to improve the effectiveness of KD~\cite{romero2014fitnets}. 
In recent years, KD has been extended to other areas. 
Goldblum~\cite{goldblum2020adversarially} analyzes the application of knowledge distillation to adversarial robustness and proposes ARD to transfer knowledge from a large teacher model with better robustness to a small student model.
ARD can produce a student network with better robustness than training from scratch. 
In this paper, we propose a more effective collaborative training framework to improve the robustness of the network.

\section{Proposed Method}
\subsection{Motivation}
\begin{figure}[t]
    \begin{center}
    \includegraphics[width=0.47\textwidth]{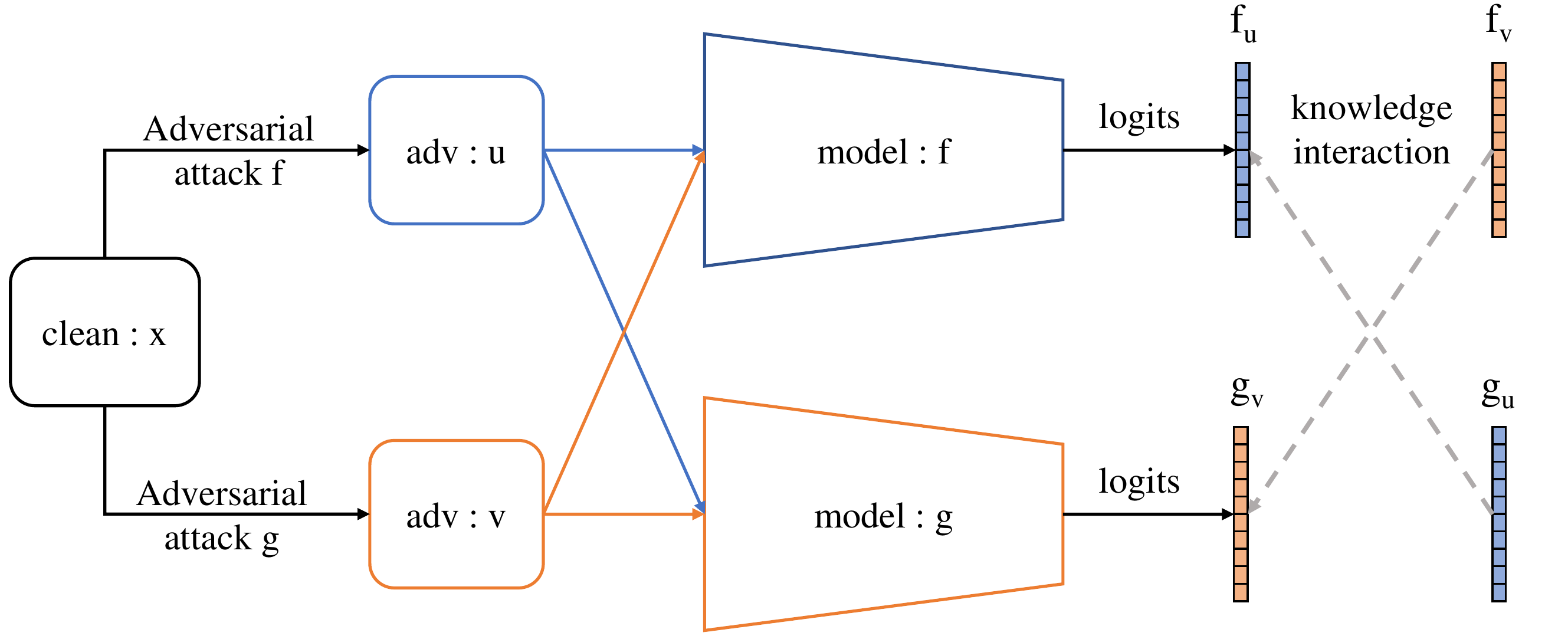}
    \end{center}
    \caption{\textbf{The framework of CAT, performing adversarial training collaboratively.}
    Given a batch of natural samples, the two networks $f$ and $g$ are attacked separately to generate adversarial examples u and v. 
    Then u and v are fed into both networks to obtain the corresponding logits. 
    We then use the logits obtained from the peer networks to guide the learning of its network, i.e., $g_u \rightarrow f_u , f_v \rightarrow g_v$.
    The process is called knowledge interaction.}
    \label{fig:framework}
\end{figure}

We investigated the characteristics of the robust models obtained by different adversarial training methods on sample instances. 
We found that different models perform differently on sample instances: for some samples, the model trained by correctly classifies AT~\cite{madry2017towards}, while the model trained by TRADES~\cite{zhang2019theoretically} misclassifies. 
Confusion exists in different methods.
A straightforward conclusion can be drawn that the networks trained by different methods master different knowledge, although their accuracy values are about the same. 
So can we use the knowledge learned from these two networks to improve the robustness of neural networks?
A simple idea is to let two networks that master different knowledge learn collaboratively. 
For this purpose, we propose collaborative adversarial training.
CAT improves the robustness of neural networks by making the knowledge of both networks interact during the training procedure.
And the framework is illustrated in \cref{fig:framework}.

\subsection{Premilinary}
\label{preliminary}
We take the methods of AT and TRADES as an example to introduce collaborative adversarial training.
We first briefly introduce the training objective functions of AT and TRADES and then introduce CAT in detail.

Adversarial training is defined as a min-maximization problem.
PGD is used to generate adversarial examples for the internal maximization process, while external minimization uses the internal PGD-generated adversarial examples and the ground-truth label $y$ to optimize the model parameters.
AT is formulated as:
\begin{equation}
    \underset{\theta}{\min} \  \mathbb{E}_{(x,y) \in D_{data}}(\underset{\delta}{\arg\max} \  L(f^{AT}_{\theta}(x^{adv}_{AT}),y)),
    \label{at-min}
\end{equation}
\begin{equation}
    x^{adv}_{AT} = x + \delta.
    \label{at-max}
\end{equation}
where $D_{data}$ is the training data distribution, $x$ and $y$ are training data and corresponding label samples from $D_{data}$. 
$f_{\theta}$ is a neural network parameterized by $\theta$. 
$L$ is the standard cross-entropy loss used in image classification tasks. 
$\delta$ is the adversarial perturbations generated by PGD.
Following previous study, $\delta$ is bounded by $l_{\infty}$.

Neural Networks trained by AT can obtain a certain level of robustness, with compromises on the accuracy of natural samples. 
For this purpose, TRADES uses a new training objective function for adversarial training. Formulated as:
\begin{equation}
\begin{split}
    \underset{\theta'}{\min} \ \mathbb{E}_{(x,y) \in D_{data}} 
    L(g^{TRADES}_{\theta'}(x), y) \quad\quad \\ +
    \lambda D_{KL}(g^{TRADES}_{\theta'}(x), g^{TRADES}_{\theta'}(x^{adv}_{TRADES})),
    \label{eq-trades}
\end{split}
\end{equation}
where $x^{adv}$ is the adversarial data corresponding to natural data $x$ and y is the true label.
$L$ is the cross-entropy loss in classification task.
$D_{KL}$ is the KL divergence to push natural logits and adversarial logits together.
The losses are balanced by a trade-off parameter $\lambda$.

\subsection{Collaborative Adversarial Training (CAT)}
\label{method}
CAT expects to improve robustness by letting neural networks trained by different methods interact with knowledge information, i.e., collaborative adversarial learning.
As illustrated in \cref{fig:framework}. we use the logit of a peer network to guide the learning of this network.
Specifically, we input the adversarial data crafted by the network trained by AT into the network trained by TRADES to get the corresponding logit.
Then use the logit obtained by the network trained by TRADES to guide the training of the network trained by AT.
The formulation goes to:
\begin{equation}
    L_1 = D_{KL}(f^{AT}(x^{adv}_{AT}), \hat{g}^{TRADES}(x^{adv}_{AT})),
    \label{trades-at}
\end{equation}
$f^{AT}$ is the network trained with AT and $g^{TRADES}$ is the network trained with TRADES.
$\hat{g}^{TRADES}(x^{adv}_{AT})$ represents that we take the logit obtained by the network trained by TRADES as constant.
$x^{adv}_{AT}$ is the adversarial data generated by $f^{AT}$ with PGD.

Similarly, to make the two networks learn collaboratively. 
We need to feed the adversarial samples generated by the TRADES network to the AT network to obtain the corresponding logit.
And then the logit obtained by the peer network is used to guide the training of the network trained by TRADES.
The loss is formulated as:
\begin{equation}
    L_2 = D_{KL}(g^{TRADES}(x^{adv}_{TRADES}) , \hat{f}^{AT}(x^{adv}_{TRADES})).
    \label{at-trades}
\end{equation}
$x^{adv}_{TRADES}$ is the adversarial example crafted by the network trained by TRADES using KL divergence function.
$\hat{f}^{AT}(x^{adv}_{TRADES})$ represents that we take the logit obtained by the network trained by AT as constant.

Empirically, models trained purely on collaborative loss will collapse.
An intuitive explanation is that the purpose of collaborative loss is to interact knowledge, but there is no knowledge when the model are lack supervision.
It is not enough to let the two networks learn from each other in this way.
Real class labels are needed to guide them. 
For this purpose, we induce supervision by combining the respective training objective functions of the two networks and the collaborative learning objective function to guide the learning of the networks together.
Therefore, the training objective function for collaborative adversarial training based on AT and TRADES is:
\begin{equation}
    L_{total} = \alpha L_{TRADES} + ( 1 - \alpha) L_2 + \alpha L_{AT} + (1 - \alpha) L_1,
    \label{cat1}
\end{equation}

where $\alpha$ is the trade-off parameter to balance the guidance of the peer network knowledge and the original objective function. 
$L_{TRADES}$ is the training objective of TRADES defined in \cref{eq-trades}. 
And $L_{AT}$ is the training objective of AT defined in \cref{at-min}. 
The first two items in \cref{cat1} are used to train model $g$ and the last two items are used to train model $f$, due to we take the peer logit as constant.

The decision boundaries learned by different adversarial training methods are different.
Under the guidance of peer network knowledge, i.e., \cref{trades-at} and \cref{at-trades}, the two networks trained by different methods continuously optimize the classification decision boundaries in the process of collaborative learning. 
Finally, both networks learn better decision boundaries than learning alone to obtain better adversarial robustness.

Our collaborative adversarial learning is a generalized adversarial training method that can be used with any two adversarial methods.
Generally, CAT can use any number of different adversarial training methods for collaborative learning. 
Results of CAT with three adversarial training methods are delayed to \cref{sec:3attack}.

\textbf{Difference with ensemble methods:}
The main difference between collaborative adversarial learning and ensemble methods is that collaborative learning involves multiple models that learn from each other, while ensemble methods involve multiple models that are combined to produce a single output. 
During testing, each model trained by collaborative learning predicts singly, since they have interacted with their knowledge during the training time.
% While ensemble methods involve multiple models all the time.
Refer to \cref{app:sec:discussion} for details.

\section{Experiment Results}
In this section, we conduct extensive experiments on popular benchmark datasets to demonstrate the effectiveness and performance of CAT.
First, we briefly introduce the experiment setup and implementation details of CAT.
Then, ablation studies are done to choose the best hyperparameters and CAT methods.
Finally, according to the best CAT methods, we report the white-box and black-box adversarial robustness on two popular benchmark datasets.

\textbf{Datasets:}
We used Three benchmark datasets, including CIFAR-10~\cite{krizhevsky2009learning}, CIFAR-100~\cite{krizhevsky2012imagenet}, and Tiny-ImageNet. 
CIFAR-10 has 10 classes. For each class, there are 5000 images for training and 1000 images for the test. 
And for CIFAR-100, there are 100 classes, and similarly, for each class, there are 500 images for training and 100 images for testing.
The image size for CIFAR-10 and CIFAR-100 is 32x32.
Tiny-ImageNet contains 100000 images of 200 classes (500 for each class) downsized to 64×64 colored images. Each class has 500 training images, 50 validation images, and 50 test images.
Three datasets are widely used for training and testing adversarial robustness.

\begin{figure*}[ht]
    \begin{center}
    \includegraphics[width=1.0\textwidth]{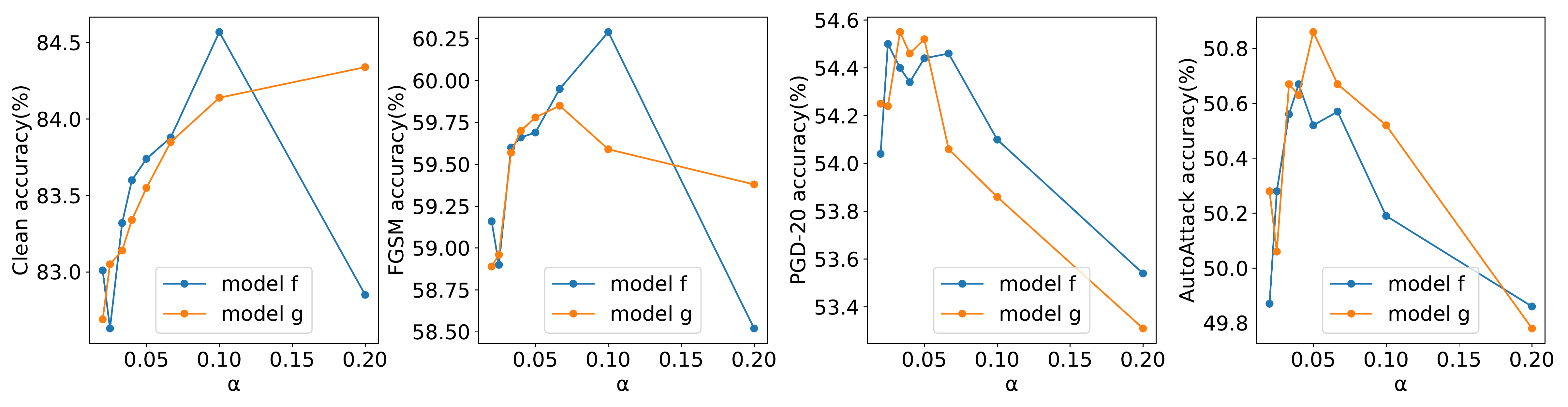}
    \end{center}
    \caption{Adversarial robustness using different hyperparameters of CAT with TRADES and AT as collaborative methods. 
    From left to right, the results of Clean acc, FGSM acc, PGD acc, and AA acc are shown.
    model $f$ and model $g$ represent the results of using TRADES and AT in the CAT training framework, respectively.}
    \label{fig:hyper} 
\end{figure*}

\begin{table*}[ht]
    \begin{center}
    \begin{tabular}{c|ccccc|ccccc}
        \toprule
        \multirow{2}{*}{Method} & \multicolumn{5}{c|}{\textbf{Best Checkpoint}}        & \multicolumn{5}{c}{\textbf{Last Checkpoint}}         \\
        & Clean & FGSM & PGD$_{20}$ & CW$_\infty$ & AA & Clean & FGSM & PGD$_{20}$ & CW$_\infty$ & AA \\ \noalign{\smallskip}\hline\noalign{\smallskip}
        \multirow{2}{*}{CAT$_{AT-TRADES}$} &83.74&	59.69	&54.44	&52.60&	50.52&	84.45&	\textbf{60.03}&	\textbf{53.01}&\textbf{52.01}&	49.30\\
        &83.55&	59.78&	\textbf{54.52}&	52.58	&50.86&	84.12&	59.69&	52.82&	51.88	&49.39\\
        \noalign{\smallskip}\hline\noalign{\smallskip}
        \multirow{2}{*}{CAT$_{AT-ALP}$} &84.66&	59.94&	53.11	&51.90	&49.74&	84.71&	59.84&	50.77&	50.53&	47.80\\
        &\textbf{85.21}	&\textbf{60.21}&	53.02&	52.13&	49.96&	\textbf{85.27}&	59.75&	51.10&	50.69&47.91\\
        \noalign{\smallskip}\hline\noalign{\smallskip}
        \multirow{2}{*}{CAT$_{TRADES-ALP}$} &83.91&	59.76	&54.44	&52.56	&\textbf{51.02}&84.67	&59.85	&52.51	&51.43	&49.31 \\
        &84.75&59.76&	54.17&	\textbf{52.72}	&50.85	&\textbf{85.27}	&59.82	&52.56	&51.83	&\textbf{49.64}\\
        \bottomrule
    \end{tabular}
    \end{center}
    \caption{The white-box robustness of different CAT methods on CIFAR-10.
    We report the results of the best checkpoint and last checkpoint. The best results are marked using \textbf{boldface}.
    ResNet-18 is the basic network in our CAT framework.}
    \label{different method}
\end{table*}

\textbf{Training setup:}
\label{train_set}
Our overall training parameters refer to~\cite{madaan2020adversarial}. 
Specifically, we use SGD (momentum 0.9, batch size 128) to train ResNet18 for 200 epochs on the CIFAR-10 dataset with weight decay 5e-4 and initial learning rate 0.1 which is divided by 10 at 100-th and 150-th epoch, respectively. 
For the internal maximization process, we use PGD$_{10}$ adversarial attack to solve, with a random start, step size 2.0/255, and perturbation size 8.0/255. 
The experimental parameters of ResNet18 in CIFAR-100, WideResNet-34-10 in CIFAR-10, and CIFAR-100 are the same as described above.

\textbf{Evaluation setup:}
\label{eval_set}
We report the clean accuracy on natural examples and the adversarial accuracy on adversarial examples. 
For adversarial accuracy, we report both white-box and black-box. 
We follow the widely used protocols in the adversarial research field. 
For the white-box attack, we consider three basic attack methods: FGSM~\cite{goodfellow2014explaining}, PGD~\cite{madry2017towards}, and CW$_{\infty}$~\cite{carlini2017towards} optimized by PGD$_{20}$, and a stronger ensemble attack method named AutoAttack (AA)~\cite{croce2020reliable}.
For the black-box attacks, we consider both transfer-based attacks and query-based attacks.

\subsection{Ablation Study}
\label{ablation}
\subsubsection{Hyperparameter:}
CAT improves adversarial robustness through the learning of collaboration, which requires both the knowledge of peer networks and the guidance of the ground truth label.
The balance of these two items is traded off by a hyperparameter $\alpha$.
We execute collaborative training by TRADES and AT as the base method and experiment with different trade-off parameters.
We test various $\alpha$ values varying from 1/50 to 1/5.
The experiment results are illustrated in \cref{fig:hyper}.
From the figure, we can conclude that if $\alpha$ is too high, i.e., little knowledge is extracted from the peer network, the effect is about the same as training with AT and trades alone. 
If $\alpha$ is too small, i.e., overly focused on the knowledge from the peer network, The network is also not very robust and will collapse when $\alpha = 0$, which is not shown in Figure.
Since Auto-Attack is currently the most powerful integrated attack method, we choose hyperparameters $\alpha$ based primarily on the robustness of the network against AA. 
In the following experiments, we set $\alpha = 0.05$ by default.

\begin{table*}[t]
    \begin{center}
    \begin{tabular}{c|c|ccccc|ccccc}
        \toprule
        \multirow{2}{*}{Dataset}&
        \multirow{2}{*}{Method} & \multicolumn{5}{c|}{\textbf{Best Checkpoint}}        & \multicolumn{5}{c}{\textbf{Last Checkpoint}}         \\
        && Clean & FGSM & PGD$_{20}$ & CW$_\infty$ & AA & Clean & FGSM & PGD$_{20}$ & CW$_\infty$ & AA \\ \noalign{\smallskip}\hline\noalign{\smallskip}
        
        \multirow{5}{*}{CIFAR-10}&
        Natural&\textbf{94.65}&19.26&0.0&0.0&0.0&\textbf{94.65}&19.26&0.0&0.0&0.0\\
        &AT &82.82&57.57&51.76&50.05&47.55&84.53&53.90&43.56&44.19&41.57 \\
        &TRADES  &83.17&59.22&52.63&50.79&49.21&83.04&57.46&49.81&	49.01&47.03\\
        &ALP &83.85	&57.20	&51.88&	50.11&	48.48&	84.64&	55.35&	44.96	&44.54	&42.62 \\
        \noalign{\smallskip}\cline{2-12}\noalign{\smallskip}
        &\multirow{2}{*}{CAT} &83.91&	\textbf{59.76}	&\textbf{54.44}	&52.56	&\textbf{51.02}	&84.67	&\textbf{59.85}	&52.51	&51.43	&49.31 \\
        &&\underline{84.75}	&\textbf{59.76}&	54.17&	\textbf{52.72}	&50.85	&\underline{85.27}	&59.82	&\textbf{52.56}	&\textbf{51.83}	&\textbf{49.64}\\
        \noalign{\smallskip}\hline\hline\noalign{\smallskip}
        
        \multirow{5}{*}{CIFAR-100}&
        Natural&\textbf{75.55}&9.48&0.0&0.0&0.0&\textbf{75.39}&9.57&0.0&0.0&0.0\\
        &AT &57.42&31.90&28.78&27.27&24.88&57.34&26.77&21.24&21.50&19.59 \\
        &TRADES  &56.98	&31.72&	29.04&	25.30&	24.23&	55.08&	30.40&	26.81	&24.78&	23.68\\
        &ALP &61.01&	31.41&	26.78&	25.68&23.51&	58.4&	27.97&	22.63&	21.87&	20.42 \\
        \noalign{\smallskip}\cline{2-12}\noalign{\smallskip}
        &\multirow{2}{*}{CAT}
        &61.31	&35.83&	\textbf{33.09}&	\textbf{29.17}&	\textbf{27.17}&	61.78&\textbf{	35.84}&	\textbf{32.76}&	\textbf{29.48}&	\textbf{27.29}\\
        &&\underline{62.53}&	\textbf{36.05}	&32.92	&29.16&	26.90&	\underline{62.52}&	35.79&	32.51&	29.24	&26.73\\
        \bottomrule
    \end{tabular}
    \end{center}
    \caption{The white-box robustness of CAT on CIFAR-10 and CIFAR-100. 
    We report the results of the best checkpoint and last checkpoint.
    ResNet-18 is the basic network in our CAT framework.}
    \label{white-ResNet-CIFAR-10-CIFAR-100}
\end{table*}

\begin{table*}[t]
    \begin{center}
    \setlength{\tabcolsep}{2.5mm}
    \begin{tabular}{c|ccccc|ccccc}
        \toprule
        \multirow{2}{*}{Method} & \multicolumn{5}{c|}{\textbf{CIFAR-10}}        & \multicolumn{5}{c}{\textbf{CIFAR-100}}         \\
         & FGSM & PGD$_{20}$&PGD$_{40}$ & CW$_\infty$ &Square & FGSM & PGD$_{20}$&PGD$_{40}$ & CW$_\infty$ & Square \\ \noalign{\smallskip}\hline\noalign{\smallskip}
        AT &64.54&61.70&61.57&61.42&56.16&39.15&37.56&37.46&38.85&30.11 \\
        TRADES&65.63&63.57&63.57&63.23&55.97&39.06&37.73&37.79&38.86&28.72\\
        ALP &64.95&62.38&62.32&61.78&55.78&40.29&38.97&38.85&40.03&29.85 \\
        \noalign{\smallskip}\hline\noalign{\smallskip}
        \multirow{2}{*}{CAT} &\underline{65.73}&\underline{63.65}&\underline{63.78}&\underline{63.24}&\underline{57.55}&\underline{42.26}&\underline{40.76}&\underline{40.76}&\underline{41.78}&\underline{33.04}\\
        &\textbf{66.06}&\textbf{63.91}&\textbf{63.88}&\textbf{63.26}&\textbf{57.95}&\textbf{42.81}&\textbf{41.55}&\textbf{41.42}&\textbf{42.42}&\textbf{33.30}\\
        \bottomrule
    \end{tabular}
    \end{center}
    \caption{The black-box robustness of CAT on CIFAR-10 and CIFAR-100. 
    We only report the results of the best checkpoint. 
    ResNet-18 is the basic network in our CAT framework.}
    \label{black-ResNet}
\end{table*}

\subsubsection{Different CAT methods:}

As described in \cref{method}, any two adversarial training methods can be incorporated into the CAT framework and learned collaboratively.
Considering that different adversarial training methods have distinct properties, the performance of different CAT methods may also vary. 
For this reason, we consider three collaborative adversarial training methods, AT-TRADES, AT-ALP, and TRADES-ALP, respectively.
The method of collaborative adversarial training using TRADES and ALP can be denoted as CAT$_{TRADES-ALP}$.
The other two methods are denoted in the same way.
Due to the fact that CAT uses two models for collaborative training, we report the results for both networks.
\cref{different method} shows the performance of CAT using different adversarial training methods. 
CAT achieves good robustness against four attack methods in all settings. 
We again mainly consider the performance of AA and choose TRADES-ALP as the base method for CAT.
Further, we analyze the correlation between discrepancy and performance after collaborative learning, which is delayed to \cref{sec:intersection}.
Without a further statement, CAT represents CAT$_{TRADES-ALP}$ in the following sections.
CAT in other settings is delayed to \cref{app:sec:1model2attack,app:sec:2model1method}

\subsection{Adversarial Robustness}
\subsubsection{White-box Robustness}
\label{white-set}
For FGSM, PGD, CW$_{\infty}$, AA, the attack perturbations are all 8.0/255 and the step size for PGD, CW$_{\infty}$ are 2/25, with 20 iterations. 
According to the way of reporting in previous papers, we report both the best checkpoint and the last checkpoint of the training phase. 
The best checkpoint result of the training phase is selected based on the model's PGD defense results for the test dataset (attack step size 2.0/255, iteration number 10, perturbation size 8.0/255).

\cref{white-ResNet-CIFAR-10-CIFAR-100} shows the adversarial accuracy of the networks trained by different methods on two datasets CIFAR-10 and CIFAR-100 against the four attacks. 
We also report the accuracy of the model for natural examples. 
From the table, we can obtain the following conclusions:
(1) our method obtains good robustness against all four attacks on both datasets. 
For example, for the strongest AA attack method, CAT can obtain 2\% improvement. 
(2) our method obtains high adversarial robustness while ensuring accuracy for natural examples.
Although there is still a big gap compared to 94.65\% of the standard training strategy, there is a nearly 1\% improvement in the accuracy of the natural examples compared to the other three methods. 
(3) The robustness of both networks is significantly improved in the CAT training framework, which is higher than separately trained ones.
(4) The difference in accuracy between the two networks trained in the CAT framework is smaller than separately trained ones, which demonstrates that the two networks do well in collaborative learning. 
For example, for the CIFAR-10 dataset, the difference in robustness between the two networks of TRADES-ALP against AA in the CAT training framework is 0.17\%, while the difference is 0.73\% under separate training. 
The same conclusion can be drawn from the results on CIFAR-100 dataset.

To investigate the generalizability of CAT with different networks, we conduct experiments with VGG-16 and MobileNet on CIFAR-10 datasets.
The results are delayed to \cref{app:sec:vgg,app:sec:mobile}.
Further, we conduct adversarial training with ResNet-18 on Tiny-ImageNet to explore the CAT on a large dataset, which is delayed to \cref{app:sec:tiny-imagenet}.
The robustness improvement holds true for all experiments.

\subsubsection{Black-box Robustness}
For black-box attacks, we consider both transfer-based attacks and query-based attacks.
For the transfer-based attack, we use the standard adversarial training of ResNet-34 as the surrogate model, trained with the same parameters as described in \cref{train_set}.
First, we perform the attack on the surrogate model to generate adversarial examples and then transfer the adversarial examples to the target network to get the robustness of the target network.
Here, we consider four attacks: FGSM, PGD$_{20}$, PGD$_{40}$, and CW$_{\infty}$, with the same attack parameters as \cref{white-set}.
For query-based attacks, we consider the Square attack, which is an efficient black-box query-based attack method.
\cref{black-ResNet} shows the results.
CAT can bring 1.79\% and 3.19\% robustness improvement against Square attack, for CIFAR-10 and CIFAR-100 respectively.
Similarly, the improvement on CIFAR-100 is more significant than CIFAR-10.

\begin{table}[t]
    \begin{center}
    \setlength{\tabcolsep}{3.0mm}
    \begin{tabular}{c|cc}
        \toprule
        Method   & Clean & AA \\
        \noalign{\smallskip}\hline\noalign{\smallskip}
        Bag of Tricks for AT~\cite{pang2020bag} &  86.28 & 53.84 \\
        HE*~\cite{pang2020boosting}  & 85.14 & 53.74 \\
        Overfitting in AT*~\cite{rice2020overfitting}&85.34 & 53.42\\
        Overfitting in AT~\cite{rice2020overfitting} &85.18 & 53.14\\
        Self-Adaptive Training~\cite{huang2020self}&83.48&53.34\\
        FAT~\cite{zhang2020attacks}&84.52&53.51\\
        TRADES~\cite{zhang2019theoretically}&84.92&53.08\\
        LLR$^{\dagger}$~\cite{qin2019adversarial}&86.28&52.84\\
        LBGAT+TRADES ($\alpha = 0$)*~\cite{cui2021learnable} & \textbf{88.70} & 53.57\\
        LBGAT+TRADES ($\alpha = 0$)~\cite{cui2021learnable}&88.22&52.86 \\
        LBGAT+TRADES ($\alpha = 6$)~\cite{cui2021learnable}&81.98&53.14 \\
        LAS-AT~\cite{las-at}&86.23&53.58\\
        LAS-AWP~\cite{las-at}&87.74&55.52\\
        \noalign{\smallskip}\hline\noalign{\smallskip}
        \multirow{2}{*}{CAT}&86.22 &\underline{54.11} \\
        &86.51 & \textbf{54.20}\\
        \midrule
        \multirow{2}{*}{CAT+AWP}&86.74 &\underline{56.43} \\
        &87.01 & \textbf{56.61}\\
        \bottomrule
    \end{tabular}
    \end{center}
    \caption{Quantitative comparison with the state-of-the-art adversarial traing methods.
    WideResNet-34-10 is the basic network in our CAT framework.
    * denotes WideResNet-34-20 network, and $\dagger$ denotes WideResNet-40-8 network.
    AWP is equipped with adversarial training methods to get better results.}
    \label{compare-to-sota}
\vspace{-0.5cm}
\end{table}

\subsection{Comparision to SOTA}

We use WideResNet-34-10~\cite{zagoruyko2016wide} networks for collaborative adversarial training to compare with previous sota methods.
\cref{compare-to-sota} shows the accuracy of the different methods for natural examples and the robustness against Auto-Attack.
From the table, we can conclude that the robustness of both networks trained with CAT outperforms the previous methods, demonstrating the state-of-the-art performance obtained by our CAT.
AWP further boosts the robustness of CAT with 2.41\% improvement.

\subsection{Comparision to KD-AT}
In general, the robustness of large models is higher than that of small models under the same training settings.
For example, WideResNet-34-10~\cite{zagoruyko2016wide} trained by TRADES can achieve 53.08\% robustness against AA, while the accuracy of ResNet-18 is only 49.21\%. 
Researchers use knowledge distillation to distill the robustness of large models to small models and obtained good results. 
We call these methods KD-AT.
Considering that CAT also involves the collaborative training of two models, we compare CAT with the KD-AT method.
To give a fair comparison, we use two different-size networks for CAT training, the same as the teacher and student network used in KD-AT.
Note that, unlike the KD method where the teacher is trained in advance, our CAT is trained with both the large model and the small model simultaneously, so there is no concept of teacher and student.
In another word, we extend previous offline distillation (2 stages) to an online way (1 stage) and achieve better performance with lower computation resources.
Illustration comparison is shown in \cref{app:sec:discussion}.

\cref{compare-to-kd} shows the results of Knowledge Distillation-AT and CAT, where ARD~\cite{goldblum2020adversarially}, IAD~\cite{zhu2021reliable}, and RSLAD~\cite{zi2021revisiting} are trained by KD-AT using TRADES trained WideResNet-34-10 network as teacher. 
CAT is collaboratively trained using two networks of different sizes. 
It can be seen that our method obtains high adversarial robustness and also obtains high clean accuracy.
More importantly, The robustness of CAT is higher than RSLAD equipped with AWP.

\begin{table}[t]
    \begin{center}
    \begin{tabular}{c|c|c|cc}
        \toprule
        Method  & Stage&Time& Clean & AA \\
        % TRADES~\cite{zhang2019theoretically} &WideResNet-34-10& 84.92 & 53.08\\
        \noalign{\smallskip}\hline\noalign{\smallskip}
        ARD~\cite{goldblum2020adversarially}&2&2720& 83.93 & 49.19\\
        IAD~\cite{zhu2021reliable}&2&2723&83.24&49.10\\
        RSLAD~\cite{zi2021revisiting}&2&2723&83.38&51.49\\
         RSLAD~\cite{zi2021revisiting}+AWP&2&-&81.26&51.62\\\noalign{\smallskip}\hline\noalign{\smallskip}
        CAT&1&2516&\textbf{84.39} & \textbf{51.72}\\
        \bottomrule
    \end{tabular}
    \end{center}
    \caption{Quantitative comparison with the state-of-the-art KD-AT methods.
    A WideResNet-34-10 and a ResNet-18 network are used in our CAT to have a fair comparison with distillation methods.
    Time denotes training time (s) per epoch.
    }
    \label{compare-to-kd}
\end{table}

\section{Property Analysis}

\subsection{Alleviate Overfitting}
\label{sec:overfit}
\begin{figure}[t]
    \begin{center}
    \includegraphics[width=0.48\textwidth]{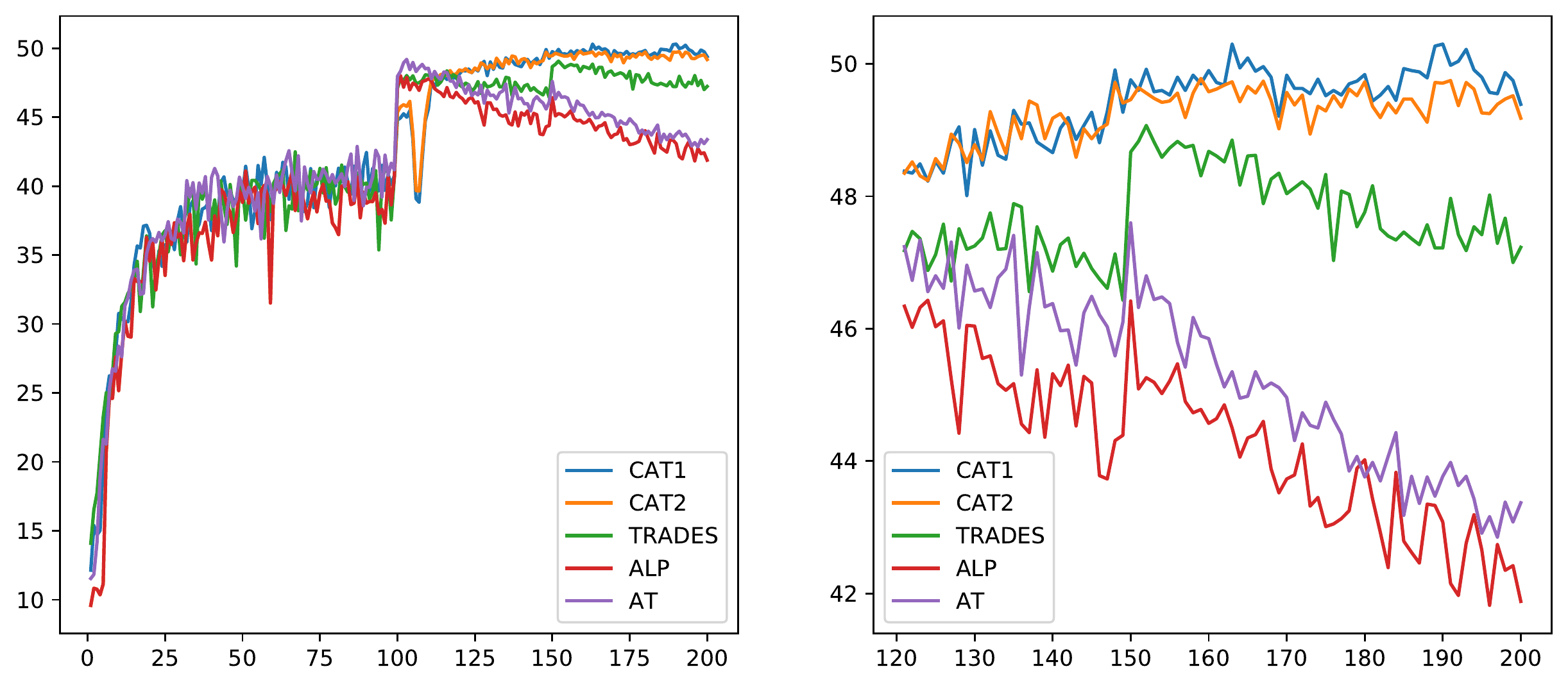}
    \end{center}
    \caption{Robust accuracy of AT, ALP, TRADES, and CAT with VGG-16 network on CIFAR-10 dataset during the adversarial training process.
    CAT can alleviate the problem of overfitting.}
    \label{fig:overfit}
\vspace{-0.2cm}
\end{figure}

Overfitting in adversarial training is first proposed by~\cite{rice2020overfitting}, which shows the test robustness decreases after peak robustness. 
And overfitting is one of the most concerning problems in adversarial training.
Here, we investigated the overfitting problem in CAT with VGG-16.
Results are illustrated in \cref{fig:overfit}.
Our CAT can alleviate the overfitting problem that widely occurs in previous adversarial methods.
Moreover, the performance for CAT has not saturated, and high performance is expected with longer epoch training.

\subsection{Correlation of discrepancy and CAT}
\label{sec:intersection}
We analyze the correlation between the discrepancy of different adversarial training methods and their adversarial robustness after CAT.
First, we compute the prediction intersection between different methods, formulated as:
\begin{equation}
    intersection = \frac{1}{N} \sum_{x_i \in D} \mathbb{I} (f^{AT}(x_i) , g^{TRADES}(x_i) ),
\end{equation}
where D is the datasets, and $\mathbb{I}$ is indicator function, which is 1 when $f^{AT}(x_i)=g^{TRADES}(x_i)$ and 0 otherwise.
Prediction discrepancy equals 1 minus intersection.
The larger this value is, the greater the discrepancy.
Then, we report the adversarial robustness of CAT trained in different settings.
Results are reported in \cref{tab:intersection}.
A conclusion can be drawn that the greater the discrepancy between different methods is, the higher the adversarial robustness after CAT.

\begin{table}[t]
    \begin{center}
    \setlength{\tabcolsep}{2.0mm}
    \begin{tabular}{c|c|c}
        \toprule
        Method& PGD$_{20}$ & Prediction discrepancy\\ \noalign{\smallskip}\hline\noalign{\smallskip}
        CAT$_{AT-ALP}$ &53.11 & 18.98\%\\
        \noalign{\smallskip}\hline\noalign{\smallskip}
        {CAT$_{TRADES-ALP}$}&54.44&21.05\%\\
        \noalign{\smallskip}\hline\noalign{\smallskip}
        CAT$_{AT-TRADES}$ &54.52 & 22.54\%\\
        \bottomrule
    \end{tabular}
    \end{center}
    \caption{The correlation between white-box robustness after CAT and prediction discrepancy of different methods on CIFAR-10.
    ResNet-18 networks are used in our CAT.}
    \label{tab:intersection}
    
\end{table}

\begin{table}[t]
    \begin{center}
    \setlength{\tabcolsep}{1.5mm}
    \begin{tabular}{c|ccccc}
        \toprule
        Method &Clean & FGSM & PGD$_{20}$ & CW$_\infty$ & AA\\ \noalign{\smallskip}\hline\noalign{\smallskip}
        
        \multirow{2}{*}{CAT$_{T-A}$} &83.91&59.76	&54.44	&52.56	&51.02 \\
        &\textbf{84.75}	&59.76&	54.17&52.72	&50.85\\
        \noalign{\smallskip}\hline\noalign{\smallskip}

        \multirow{3}{*}{CAT$_{A-A-T}$}
        &84.50 & 60.17 & 54.64 & 52.98 & 51.28\\
        &84.62 & \textbf{60.25 }& 54.87 & 53.04 & 51.42\\
        &84.29 & 60.24 & \textbf{55.04} & \textbf{53.38} & \textbf{51.74}\\
        \bottomrule
    \end{tabular}
    \end{center}
    \caption{The white-box robustness of CAT on CIFAR-10. 
    We report the results of the best checkpoint.
    Resnet-18 is the basic network of CAT.
    T-A is short for TRADES-ALP, denoting collaboration between TRADES and ALP.
    A-A-T is short for AT-ALP-TRADES, denoting collaboration between AT, ALP, and TRADES.}
    \label{white-ResNet-baseline}
\end{table}

\subsection{CAT of Three models with three methods}
\label{sec:3attack}
Our CAT is a generalized method, which can use any number of different adversarial training methods for collaborative learning.
We conduct an experiment on CAT by collaborating three adversarial training methods.
The results are reported in \cref{white-ResNet-baseline}.
The robustness improvement is more significant than CAT trained with two adversarial-trained methods, which shows the generalizability of our CAT. 
Collaborating three methods can bring 0.7\% improvement against Auto-Attack.

\section{Conclusion}
\label{conclusion}
In this paper, we first analyze the properties of different adversarial training methods and find that networks trained by different methods perform differently on sample instances, i.e., the network can correctly classify samples that are misclassified by other networks. 
Based on this observation, we propose a collaborative adversarial training framework to improve the robustness of both networks.
CAT aims to guide network learning using true label supervision together with the knowledge mastered in peer networks, which is different from its own knowledge.
Extensive experiments on different datasets and different networks demonstrate the effectiveness of our approach, and state-of-the-art performance is achieved. 
Furthermore, property analysis is conducted to get a better understanding of CAT.
Broadly, CAT can be easily extended to multiple networks for collaborative adversarial training. 
We hope that CAT brings a new perspective to the study of adversarial training.

{\small
\bibliographystyle{ieee_fullname}
\bibliography{egbib}
}

%--------------------------appendix----------------

\clearpage
\newpage
\appendix

\renewcommand\thesection{\Alph{section}} 
\renewcommand\thesubsection{\Alph{section}.\arabic{subsection}} 
\renewcommand\thefigure{\Alph{section}\arabic{figure}} 
\renewcommand\thetable{\Alph{section}\arabic{table}} 
\setcounter{section}{0}
\setcounter{figure}{0}	
\setcounter{table}{0}

\section{More experimental results}

\subsection{VGG-16 results on CIFAR-10}
\label{app:sec:vgg}
The white-box robustness of VGG-16~\cite{vgg} models trained using AT, ALP, TRADES, and CAT are reported in \cref{white-vgg-CIFAR-10}.
The setting for VGG-16 is the same as ResNet-18 models, i.e., $\alpha = 1.0/20$ and $\beta = 1.0/20$.
The improvement for CAT with VGG-16 models is as consistent with ResNet-18 models.
CAT can boost model's robustness under AutoAttack with 2.0 points.

\subsection{Mobilenet results on CIFAR-10}

Similar to the above VGG-16 models, we report the while-box robustness of MobileNet~\cite{mobilenet} on CIFAR-10 datasets under various attacks in \cref{white-mobilenet-CIFAR-10}.
The experiment set is the same as the previous setting.
We can see that our CAT brings 1.0 improvement for MobileNet under AutoAttack, which is the most powerful adversarial attack method.

\subsection{ResNet-18 results on Tiny-ImageNet}
For the large-scale ImageNet dataset, just as all the baseline methods did not report the results, we are also unable to evaluate on ImageNet due to the very high training cost.
To investigate the performance of our CAT in large datasets, we conduct the experiment of white-box robustness of ResNet-18 on Tiny-ImageNet, which also is a widely used dataset in adversarial training.
The results are shown in \cref{white-resnet-tiny-imagenet}.
Surprisingly, CAT shows impressive robustness on the large-scale dataset.
The improvement is as significant as ResNet-18 in small datasets like CIFAR-10 and CIFAR-100.

\subsection{CAT of One model with various attacks}
\label{app:sec:1model2attack}
For our CAT method, we use two networks and two different attack methods for each network to perform adversarial training.
An interesting baseline is one network with two different attack methods.
Therefore, we use PGD and CW as our attack methods and one ResNet-18 as our network.
The results are reported in \cref{white-ResNet-baseline} (CAT$_{P-C}$ entry).
The improvement for this setting is not significant as the previous setting, but it still, boosts the model's robustness against all four attacks.

\subsection{CAT of Two models with same methods}
\label{app:sec:2model1method}
Another interesting baseline is two networks trained by the same adversarial training methods, i.e., two ResNet-18 networks are both trained by TRADES.
We denote this setting as CAT$_{T-T}$.
The results are reported in \cref{white-ResNet-baseline}.
The improvement for this setting is not significant as the previous setting, but it still, boosts the model's robustness against all four attacks.
However, the improvement is more significant than just using one network.
A conclusion can be drawn that two networks are important for CAT to achieve better adversarial robustness.

\begin{table}[t]
    \begin{center}
    \begin{tabular}{c|ccccc}
        \toprule
        Method& Clean & FGSM & PGD$_{20}$ & CW$_\infty$ & AA \\ \noalign{\smallskip}\hline\noalign{\smallskip}
        AT &78.31&53.11&48.39&46.32&43.69 \\
        TRADES  &79.11 & 53.75 & 48.28 & 45.93 & 44.63\\
        ALP & 80.23 & 52.18 & 47.30 & 45.23 & 43.68 \\
        \noalign{\smallskip}\cline{1-6}\noalign{\smallskip}
        \multirow{2}{*}{CAT}&79.23 & 54.47 & \textbf{49.43} & 47.19 & \textbf{45.48}\\
        &80.12 & \textbf{54.48} & 48.30 & \textbf{47.23} & 45.33\\
        \bottomrule
    \end{tabular}\end{center}
        \caption{The white-box robustness results (accuracy $(\%)$) of CAT on CIFAR-10.
    We report the results of the best checkpoint. The best results are marked using \textbf{boldface}. 
    \textbf{Two VGG-16} networks are used in our CAT framework.}
    \label{white-vgg-CIFAR-10}
    
\end{table}

\label{app:sec:mobile}
\begin{table}[t]
    \begin{center}
    \begin{tabular}{c|ccccc}
        \toprule
        Method& Clean & FGSM & PGD$_{20}$ & CW$_\infty$ & AA \\ \noalign{\smallskip}\hline\noalign{\smallskip}
        AT &76.24 & 50.27 & 44.99 & 43.03& 40.10\\
        TRADES  & 75.84 & 49.65 & 45.26 & 42.04 & 41.08\\
        ALP &79.46&50.14&43.95&42.08&40.01 \\
        \noalign{\smallskip}\hline\noalign{\smallskip}
        \multirow{2}{*}{CAT} &80.14&51.25&\textbf{46.38}&\textbf{44.24}&\textbf{42.20}\\
        &79.86&\textbf{51.28}&46.22&44.05&42.16\\
        \bottomrule
    \end{tabular} \end{center}
    \caption{The white-box robustness results (accuracy $(\%)$) of CAT on CIFAR-10.
    We report the results of the best checkpoint. The best results are marked using \textbf{boldface}. 
    \textbf{Two MobileNet} networks are used in our CAT framework.}
    \label{white-mobilenet-CIFAR-10}
   
\end{table}

\label{app:sec:tiny-imagenet}
\begin{table}[t]
    \begin{center}
    \setlength{\tabcolsep}{3.0mm}
    \begin{tabular}{c|cccc}
        \toprule
        Method& Clean& PGD$_{50}$ & CW$_\infty$ & AA \\ \noalign{\smallskip}\hline\noalign{\smallskip}
        AT &43.98 & 19.98 & 17.60 & 13.78 \\
        TRADES  &39.16 & 15.74 & 12.92 & 12.32\\
        ALP & 39.85 & 17.28 &15.34 & 12.98\\
        \noalign{\smallskip}\hline\noalign{\smallskip}
        \multirow{2}{*}{CAT} 
        &44.35 & 20.86 & 19.43 & 14.96\\
        &\textbf{44.76} & \textbf{21.02} & \textbf{19.64}& \textbf{15.63}\\
        \bottomrule
    \end{tabular}\end{center}
        \caption{The white-box robustness results (accuracy $(\%)$) of CAT on \textbf{Tiny-ImageNet}.
    We report the results of the best checkpoint. The best results are marked using \textbf{boldface}. 
    Two ResNet-18 networks are used in our CAT framework.}
    \label{white-resnet-tiny-imagenet}
    
\end{table}

\begin{table}[t]
    \begin{center}
    \begin{tabular}{c|ccccc}
        \toprule
        \multirow{2}{*}{Method} & \multicolumn{5}{c}{\textbf{Best Checkpoint}}         \\
        & Clean & FGSM & PGD$_{20}$ & CW$_\infty$ & AA\\ \noalign{\smallskip}\hline\noalign{\smallskip}
        AT &82.82&57.57&51.76&50.05&47.55\\
        \noalign{\smallskip}\hline\noalign{\smallskip}

        \multirow{2}{*}{CAT$_{T-A}$} &83.91&\textbf{59.76}	&\textbf{54.44}	&52.56	&\textbf{51.02} \\
        &\textbf{84.75}	&\textbf{59.76}&	54.17&\textbf{52.72}	&50.85\\
        \noalign{\smallskip}\hline\noalign{\smallskip}
        
        CAT$_{P-C}$&82.09 &56.48 &52.48 & 49.28 & 48.06 \\
        \noalign{\smallskip}\hline\noalign{\smallskip}

        \multirow{2}{*}{CAT$_{T-T}$}
        &81.94 &58.85 &54.19&51.52& 50.30\\
        &82.13&58.77&54.02&51.56& 50.14\\
        \bottomrule
    \end{tabular}\end{center}
    \caption{The white-box robustness results (accuracy $(\%)$) of CAT on CIFAR-10. 
    We report the results of the best checkpoint. The best results are marked using \textbf{boldface}.
    P-C denotes one network trained by PGD and CW.
    T-A is short for TRADES-ALP, denoting two networks with TRADES and ALP.
    T-T is short for TRADES-TRADES, denoting two networks with TRADES and TRADES.}
    \label{white-ResNet-baseline}

\end{table}

\section{Discussion}
\label{app:sec:discussion}

\begin{figure}[t]
    \begin{center}
    \includegraphics[width=0.47\textwidth]{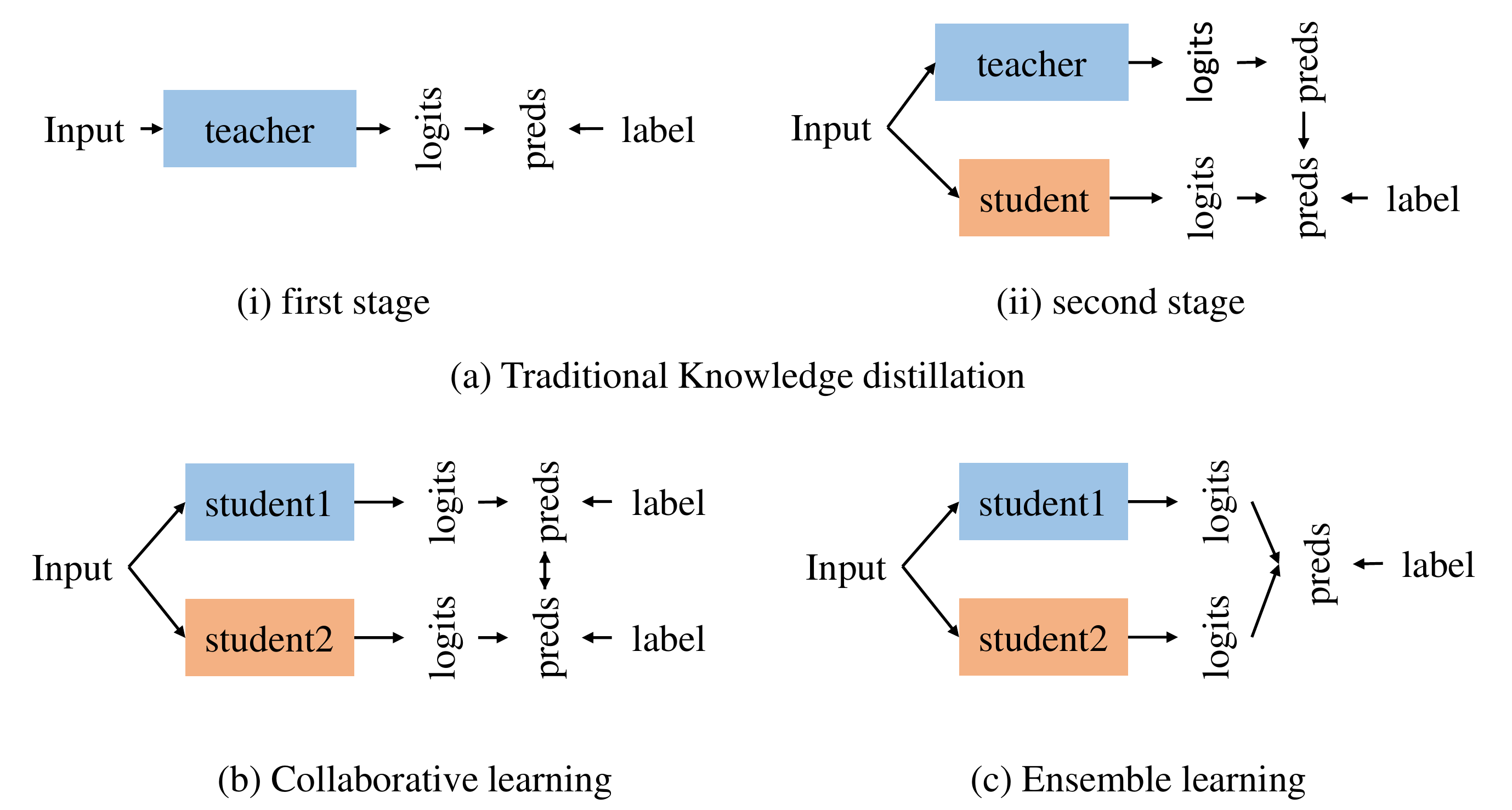}
    \end{center}
    \caption{Three types of distillation.
    (a) displays traditional knowledge distillation, which involves two-stage optimization and a large-scale teacher model.
    (b) and (c) illustrate online learning, \ie, collaborative learning and ensemble learning, which do not involve teacher models.}
    \label{app:fig:diff}
\end{figure}

In this section, we illustrate three types of distillation methods, shown in \cref{app:fig:diff}.
Traditional knowledge distillation has a two-stage optimization, which is pre-training the large-scale teacher model and distilling students with pre-trained teachers in the first stage.
RSLAD~\cite{zi2021revisiting} is implemented in this paradigm.
Two-stage optimization brings a large computation cost.
Compared to RSLAD, our CAT is based on collaborative learning and only needs one-stage optimization with two student models.

\end{document}